\documentclass[a4paper]{llncs}

\usepackage{amssymb}
\usepackage{graphicx}
\usepackage{syntax}
\usepackage{amsmath}
\usepackage{subcaption}
\usepackage{algorithm}
\usepackage{algorithmicx}
\usepackage{algpseudocode}
\usepackage{caption}

\usepackage{url}
\urldef{\mailsa}\path|jessicac@student.dei.uc.pt,{naml,penousal}@dei.uc.pt|    
\newcommand{\keywords}[1]{\par\addvspace\baselineskip
\noindent\keywordname\enspace\ignorespaces#1}
\begin{document}

\mainmatter  

\title{Probabilistic Grammatical Evolution}

\titlerunning{Probabilistic Grammatical Evolution}

%
%
\author{Jessica M\'egane\and Nuno Lourenço\and Penousal Machado}
\authorrunning{Probabilistic Grammatical Evolution}

\institute{CISUC, Department of Informatics Engineering, University of Coimbra,\\Polo II - Pinhal de Marrocos, 3030 Coimbra, Portugal\\ \mailsa}

%
%

\toctitle{Lecture Notes in Computer Science}
\tocauthor{Authors' Instructions}
\maketitle

\begin{abstract}
Grammatical Evolution (GE) is one of the most popular Genetic Programming (GP) variants, and it has been used with success in several problem domains. Since the original proposal, many enhancements have been proposed to GE in order to address some of its main issues and improve its performance. 

In this paper we propose Probabilistic Grammatical Evolution (PGE), which introduces a new genotypic representation and new mapping mechanism for GE. Specifically, we resort to a Probabilistic Context-Free Grammar (PCFG) where its probabilities are adapted during the evolutionary process, taking into account the productions chosen to construct the fittest individual. The genotype is a list of real values, where each value represents the likelihood of selecting a derivation rule. We evaluate the performance of PGE in two regression problems and compare it with GE and Structured Grammatical Evolution (SGE). 

The results show that PGE has a a better performance than GE, with statistically significant differences, and achieved similar performance when comparing with SGE.

\keywords{Genetic Programming, Grammatical Evolution, Probabilistic Context-Free Grammar, Probabilistic Grammatical Evolution, Genotype-to-Phenotype Mapping}
\end{abstract}

\section{Introduction}
Evolutionary Algorithms (EAs) are loosely inspired by the ideas of natural evolution, where a population of individuals evolves through the application of selection, variation (such as crossover and mutation) and reproduction operators. The evolution of these individuals is guided by a fitness function, which measures the quality of the solutions that each individual represents to the problem at hand. The application of these elements is repeated for several iterations and it is expected that, over time, the quality of individuals improves.

Genetic Programming (GP) \cite{koza} is an EA that is used to evolve programs. Over the years many variants of GP have been proposed, namely concerned with how individuals (i.e., computer programs) are represented. Some of these variants make use of grammars to enforce syntactic constraints on the individual solutions. The most well known grammar-based GP approaches are Context-free Grammar Genetic Programming (CFG-GP), introduced by Whigham in \cite{whigham}, and Grammatical Evolution (GE) introduced by Ryan \textit{et al.} \cite{oneill98}. The main distinction between the two approaches is the representation of the individual’s solution (genotype) in the search space. CFG-GP uses a derivation-tree based representation, and the mapping is made by reading the terminal symbols (tree leaves), starting from the left leaf to the right. In GE there is a distinction between the genotype, a variable length string of integers, and the phenotype of the individual. The mapping between the genotype and the phenotype is performed through a Context-Free Grammar (CFG). 

GE is one of the most popular GP variants, in spite of the debate in the literature \cite{Whigham2015} concerning its relative performance when compared to other grammar-based variants. To address some of the main criticisms of GE, several improvements have been proposed in the literature related to the population initialisation \cite{Nicolau2017}, grammar design \cite{Nicolau2018} and the representation of individuals \cite{ONeill2004,Kim2015,KIM2016,Loureno2016}.

In this paper we introduce a new representation to GE. In concrete, we proposed a new probabilistic mapping mechanism to GE, called Probabilistic Grammatical Evolution (PGE). In PGE the genotype is a list of probabilities and the mapping is made using a Probabilistic Context-Free Grammar (PCFG) to choose the productions of the individual's phenotype. All derivation rules in the grammar start with the same chance of being selected, but over the evolutionary process, the probabilities are updated considering the derivation rules that were selected to build the fittest individual. 
To evaluate the performance of PGE, we compare its performance with GE and SGE \cite{Loureno2017} in two benchmark problems. PGE showed statistically significant improvements when compared with GE and obtained similar performance when compared to SGE.

The remainder of the paper is structured as follows: Section \ref{sec:ge} introduces Grammatical Evolution and related work. Section \ref{sec:pge} describes our approach called Probabilistic Grammatical Evolution (PGE), Section \ref{sec:ea} details the experimental framework used to study the performance of PGE, and Section \ref{sec:results} describes the main results. Finally, Section \ref{sec:conclusion} gathers the main conclusions and provides some insights towards future work.

\section{Grammatical Evolution}
\label{sec:ge}
GE \cite{oneill98} is a Grammar-based GP approach where the individuals are presented as a variable length string of integers. To create an executable program, the genotype (i.e., the string of integers) is mapped to the phenotype (program) via the productions rules defined in a Context-Free Grammar (CFG). A grammar is a tuple $G = (NT,T,S,P)$ where $NT$ and $T$ represent the non-empty set of \textit{Non-Terminal} (NT) and \textit{Terminal} (T) symbols, $S$ is an element of $NT$ called the axiom and $P$ is the set of production rules. The rules in $P$ are in the form $A ::= \alpha$, with $A \in NT$ and $\alpha \in (NT \cup T)^*$. The $NT$ and $T$ sets are disjoint. Each grammar defines a language $L(G)= \{ w:\, S\overset{*} {\Rightarrow} w,\, w \in T^*\}$, that is the set of all sequences of terminal symbols that can be derived from the axiom.

\begin{figure}[ht]
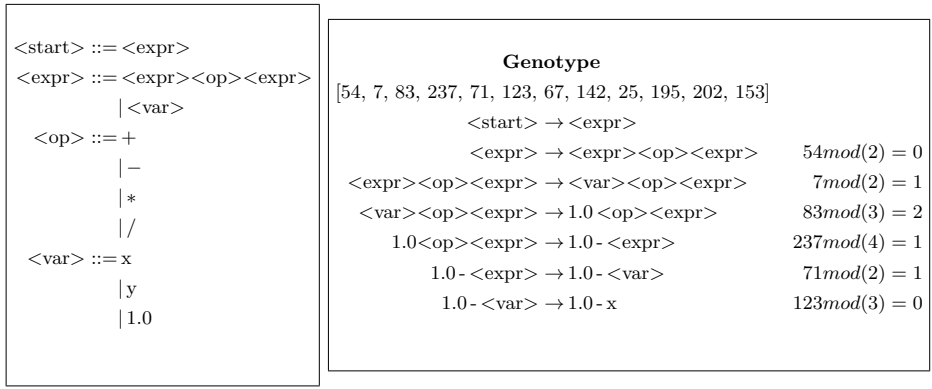

\centering
\noindent\scalebox{0.81}{\fbox{%
    \parbox{0.1\textwidth}{%
\begin{align*}
{<}\text{start}{>} ::= & \, {<}\text{expr}{>} \\
{<}\text{expr}{>} ::=  & \, {<}\text{expr}{>}{<}\text{op}{>}{<}\text{expr}{>} \\
            & | \, {<}\text{var}{>} \\
{<}\text{op}{>} ::= & \,+ \, \\
            & | \, - \\
            & | \, * \\
            & | \, / \\
{<}\text{var}{>} ::= &  \, \text{x} \, \\ 
            & | \, \text{y} \,  \\
             & | \, \text{1.0} \,  \\
\end{align*}}}}
\scalebox{0.81}{\fbox{%
    \parbox{0.1\textwidth}{%
\begin{align*}
\textbf{Genot} & \textbf{ype}\\
{[}\text{54, 7, 83, 237, 71, 123, 67} & \text{, 142, 25, 195, 202, 153}{]}\\
{<}\text{start}{>} \rightarrow & \, {<}\text{expr}{>} & \\
{<}\text{expr}{>} \rightarrow  & \, {<}\text{expr}{>}{<}\text{op}{>}{<}\text{expr}{>} & 54 mod(2) = 0 \\
{<}\text{expr}{>}{<}\text{op}{>}{<}\text{expr}{>} \rightarrow & \, {<}\text{var}{>}{<}\text{op}{>}{<}\text{expr}{>} & 7 mod(2) = 1 \\
{<}\text{var}{>}{<}\text{op}{>}{<}\text{expr}{>} \rightarrow & \, \text{1.0} \, {<}\text{op}{>}{<}\text{expr}{>} & 83 mod(3) = 2\\
\text{1.0}{<}\text{op}{>}{<}\text{expr}{>} \rightarrow & \, \text{1.0} \, \text{-} \, {<}\text{expr}{>} & 237 mod(4) = 1\\
\text{1.0} \, \text{-} \, {<}\text{expr}{>} \rightarrow & \, \text{1.0} \, \text{-} \, {<}\text{var}{>} & 71 mod(2) = 1 \\
\text{1.0} \, \text{-} \, {<}\text{var}{>} \rightarrow & \, \text{1.0} \, \text{-} \, \text{x} & 123 mod(3) = 0 \\
\end{align*}}}}

\caption{\label{fig:gemapping}Example of GE mapping}
\end{figure}

The genotype-phenotype mapping is the key issue in GE, and it is performed in several successive steps. To select which derivation rule should be selected to replace a NT, the mapping relies on the modulo operator. An example of this process is shown in Fig. \ref{fig:gemapping}. The genotype is composed of integers values randomly generated between [0,255]. The mapping starts with the axiom $<start>$. In this case, there is only one derivation possible, and we rewrite the axiom with $<expr>$. Then we proceed the expansion of $<expr>$. Since this NT has two possible expansion rules available, we need to select which one will be used. We start by taking the first unused value of the genotype, which is 54, and divide it by the number of possible options. The remainder of this operation indicates the option that should be used. In our example, $54 mod(2) = 0$, which results in the first production being selected. This process is performed until there are no more NT symbols to expand or there are no more integers to read from the genotype. 

In this last case and if we still have NT to expand, a wrapping mechanism can be used, where the genotype will be re-used, until it generates a valid individual or the predefined number of wraps is over. If after all the wraps we still have not mapped all the NT, the mapping process stops, and the individual will be considered invalid.


Even though GE has been applied to several problem domains, there is a debate in the literature concerned with its overall performance \cite{Whigham2015,Ryan2017}. GE has been criticised for having high redundancy and low locality \cite{keijzer02,Rothlauf2006}. A representation has high redundancy when several different genotypes correspond to one phenotype. Locality is concerned with how changes in the genotype are reflected on the phenotype. 
These criticisms have triggered many researchers into looking how GE could be improved \cite{chorus,ONeill2004,Bartoli2018,Loureno2017}.

\subsection{Representation Variants} 

O'Neill \textit{et al.} \cite{ONeill2004} proposed Position Independent GE ($\pi$GE), introducing a different mapping mechanism that removes the positional dependency that exists in GE. In $\pi$GE each codon is composed of two values (nont, rule), where nont is used to select the next non-terminal to be expanded and the rule selects which production rule should be applied from the selected non-terminal. In Fagan \textit{et al.} \cite{Fagan2010} several different mapping mechanisms where compared, and $\pi$GE showed better performance over GE, with statistical differences.
Another attempt to make GE position independent is Chorus \cite{chorus}. In this variant, each gene encodes a specific grammar production rule, not taking into consideration the position. This proposal did not showed significant differences when comparing with standard GE.

Structured Grammatical Evolution is a recent proposal to address the locality and redundancy problems of GE \cite{Loureno2017,Loureno2016}. SGE proposes a one-to-one mapping between the genotype and the non-terminal symbols of the grammar. Each position in the genotype of SGE is a list of integers, where each element of this list is used to select the next derivation rule. This genotype structure, ensures that the modification of a codon does not affect the chosen productions of other non-terminals, reducing the overall changes that can occur at the phenotypical level, which results in a higher locality. 

In \cite{Loureno2017} different grammar-based GP approaches were compared, and the authors showed that SGE achieved a good performance when compared with several grammar-based GP representations. These results were in line with Fagan \textit{et al.} \cite{Fagan2010}, which showed that different genotype-phenotype mapping can improve the performance of grammar-based GP.

Some probabilistic methods have been proposed to try to understand more about the distribution of fitter individuals and have been effective in solving complex problems \cite{Kim2013}. Despite its potential, few attempts have been made to use probabilities in GE. 

In \cite{Kim2015}, was implemented a PCFG (Figure \ref{fig:pcfg}) to do the mapping process of GE, where the genotype of the individual is a vector of probabilities used to choose the productions rules.
This approach also implements Estimation of Distribution Algorithms (EDA) \cite{EDA}, a probabilistic technique that replaces the mutation and crossover operators, by sampling the probability distribution of the best individuals, to generate the new population, each generation. The probabilities start all equal and are updated each generation, based on the frequency of the chosen rules of the individuals with higher fitness. 
The experiments were inclusive, since the proposed approach had a similar performance to GE. 

Kim \textit{et al.} \cite{KIM2016} proposed Probabilistic Model Building Grammatical Evolution (PMBGE), which utilises a Conditional Dependency Tree (CDT) that represents the relationships between production rules used to calculate the new probabilities. Similar to \cite{Kim2015}, the EDA mechanism was implemented instead of the genetic operators. The results showed no statistical differences between GE and the proposed approach.

\section{Probabilistic Grammatical Evolution}
\label{sec:pge}
Probabilistic Grammatical Evolution (PGE) is a new representation for Grammatical Evolution. In PGE we rely on a Probabilistic Context-Free Grammar (PCFG) to perform the genotype-phenotype mapping. A PCFG is a quintuple $PG = (NT,T,S,P, Probs)$ where NT and T represent the non-empty set of \textit{Non-Terminal} (NT) and \textit{Terminal} (T) symbols, respectively, $S$ is an element of $NT$ called the axiom, $P$ is the set of production rules, and $Probs$ is a set of probabilities associated with each production rule. The genotype in PGE is a vector of floats, where each element corresponds to the probability of choosing a certain derivation rule. The overall mapping procedure is shown in Alg. \ref{mappingPGE} and Fig. \ref{fig:pcfg} shows an example of the application of the PGE mapping.

The panel on the left shows a PCFG, where each derivation rule has a probability associated. The set of NT is composed of $<start>, <expr>, <op>$ and $<var>$. The right panel of Fig. \ref{fig:pcfg} shows how the mapping procedure works.

\begin{figure}[ht]
\centering
\noindent\scalebox{0.82}{\fbox{%
    \parbox{0.1\textwidth}{%
\begin{align*}
{<}\text{start}{>} ::= & \, {<}\text{expr}{>} & (1.0)\\
{<}\text{expr}{>} ::=  & \, {<}\text{expr}{>}{<}\text{op}{>}{<}\text{expr}{>} & (0.5)\\
            & | \, {<}\text{var}{>} & (0.5)\\
{<}\text{op}{>} ::= & \,+ \, & (0.33)\\
            & | \, * & (0.33)\\
            & | \, - & (0.33)\\
{<}\text{var}{>} ::= &  \, \text{x} \, & (0.5) \\ 
            & | \, \text{1.0} \, & (0.5) \\
\end{align*}}}}
\scalebox{0.82}{\fbox{%
    \parbox{0.1\textwidth}{%
\begin{align*}
\textbf{Genot} & \textbf{ype}\\
{[}\text{0.8, 0.2, 0.98, 0.45} & \text{, 0.62, 0.37, 0.19}{]}\\ \\
{<}\text{start}{>} \rightarrow & \, {<}\text{expr}{>} & (0.8)\\
{<}\text{expr}{>} \rightarrow  & \, {<}\text{expr}{>}{<}\text{op}{>}{<}\text{expr}{>} & (0.2) \\
{<}\text{expr}{>}{<}\text{op}{>}{<}\text{expr}{>} \rightarrow & \, {<}\text{var}{>}{<}\text{op}{>}{<}\text{expr}{>} & (0.98)\\
{<}\text{var}{>}{<}\text{op}{>}{<}\text{expr}{>} \rightarrow & \, \text{x} \, {<}\text{op}{>}{<}\text{expr}{>} & (0.45)\\
\text{x}{<}\text{op}{>}{<}\text{expr}{>} \rightarrow & \, \text{x} \, \text{*} \, {<}\text{expr}{>} & (0.62)\\
\text{x} \, \text{*} \, {<}\text{expr}{>} \rightarrow & \, \text{x} \, \text{*} \, {<}\text{var}{>} & (0.37) \\
\text{x} \, \text{*} \, {<}\text{var}{>} \rightarrow & \, \text{x} \, \text{*} \, \text{x} & (0.19) \\
\end{align*}}}}

\caption{\label{fig:pcfg}Example of mapping with PCFG}
\end{figure}

\begin{algorithm}
\caption{Mapping with PCFG}
\begin{algorithmic}[1]
\Procedure{generateIndividual}{$genotype, pcfg$}
\State start = pcfg.getStart()
\State phenotype = [start]
\For{codon in genotype}
\State symbol = phenotype.getNextNT()
\State productions = pcfg.getRulesNT(symbol)
\State cum_prob = 0.0 \Comment{Cumulative Sum of Probabilities} 
\For{prod in productions}
\State cum_prob = cum_prob + prod.getProb()
\If{codon $<$ cum_prob}
\State selected_rule = prod
\State break
\EndIf
\EndFor
\State phenotype.replace(symbol, selected_rule)
\If{phenotype.isValid()}
\State break
\EndIf
\EndFor
\EndProcedure
\end{algorithmic}
\label{mappingPGE}
\end{algorithm}

It begins with the axiom, $<start>$. We start by taking the first value of the genotype, which is 0.8, and since there is only one expansion available, the non-terminal $<expr>$ will be chosen. Next, we need to rewrite $<expr>$, which has two derivation options. We take the second value of the genotype, 0.2, and compare it to the probability associated with the first derivation option ( $<expr><op><expr>$). Since $0.2 < 0.5$, we select this derivation option to rewrite  $<expr>$. This process is repeated until there are no more non-terminals left to expand, or no probabilities left in the genotype. When this last situation occurs, we use a wrapping mechanism similar to the standard GE, where the genotype will be reused a certain number of times. If after the wrapping we still have not mapped the individual completely, the mapping process stops, and the individual will be considered invalid. 


In PGE, the probabilities are updated each generation after evaluating the population, based on how many times each derivation rule has been selected by the best individual of the current generation or the best individual overall. When a derivation rule is used to create one of these individuals, its probability should be increased, otherwise if a derivation rule is not used, we should decrease it. Alternating between these two bests helps us to avoid using the same individual in consecutive generations to adjust the probabilities of the PCFG, balancing global exploration with local exploitation. All the adjustments are performed using a parameter $\lambda$ called \textit{learning factor} which smooths the transitions on the search space. The lambda value should be between 0\% and 100\%. At each generation, each individual is mapped using an updated version of grammar.

To update the probabilities in the grammar, we use Alg. \ref{algo}, where \textit{j} is the number of productions of a non-terminal symbol of the grammar, \textit{i} is the index of the production probability that is being updated and $\lambda$ is the learning factor.

\begin{algorithm}
\caption{Probabilistic Grammatical Evolution}\label{eqpge}
\begin{algorithmic}[1]
\Procedure{updateProbabilities}{$best$}
\State $counter = best.getCounter()$    \Comment{list with times each rule was expanded}
\For{each production rule \textit{i} of each NT}
\If{$counter_i >$ 0}
\State $prob_i = min( prob_i + \frac{\lambda * counter_i}{\sum_{k=1}^{j} counter_k} , 1.0)$
\Else
\State $prob_i = prob_i - \lambda * prob_i $
\EndIf
\EndFor
\While{$\sum_{k=1}^{j} prob_i \neq 1.0$}
\State extra = (1.0 - $\sum_{k=1}^{j} prob_i$) / j
\For{each production rule \textit{i}}
\State $prob_i = prob_i + extra$
\EndFor
\EndWhile
\EndProcedure
\end{algorithmic}
\label{algo}
\end{algorithm}

The probabilities are updated based on two different rules. The first rule increases the probability of a derivation rule, taken into account the frequency that it was selected by the best individual (Alg. \ref{algo} line 5). The second rule decreases the probability of the derivation options that are never used to expand a non-terminal (Alg. \ref{algo} line 7). The $min$ operator ensures that when we update the probabilities they are not greater than 1.

After the update of the probabilities for each derivation rule, we make sure that the sum of the probabilities of all derivation rules, for a non-terminal, is 1. If the sum surpasses this value, the excess is proportionally subtracted from the derivation options for a non-terminal. If the sum is smaller than one, the missing amount is added equally to the production rules of the non-terminal.

\begin{figure}[ht]
\centering
\noindent
\scalebox{0.14}{\fbox{%
    \parbox{0.1\textwidth}{%
\centering
\begin{align*}
\includegraphics{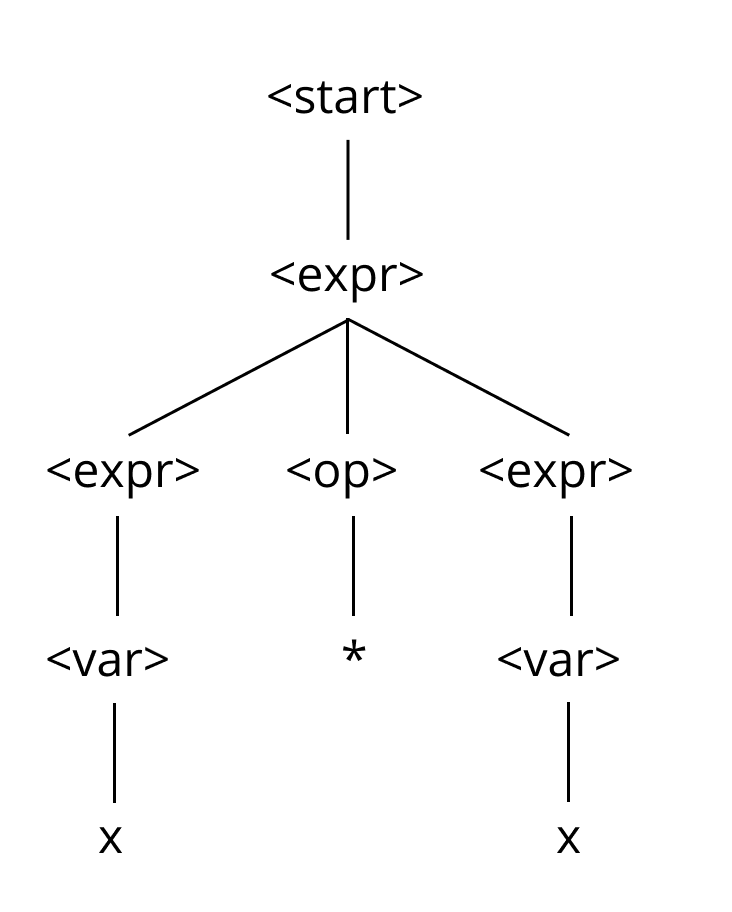}
\end{align*}}}}
\scalebox{0.82}{\fbox{%
    \parbox{0.1\textwidth}{%
\begin{align*}
\centering
\textbf{New } & \textbf{Grammar} \\
{<}\text{start}{>} ::= & \, {<}\text{expr}{>} & (1.0)\\
{<}\text{expr}{>} ::=  & \, {<}\text{expr}{>}{<}\text{op}{>}{<}\text{expr}{>} & (0.498)\\
            & | \, {<}\text{var}{>} & (0.502)\\
{<}\text{op}{>} ::= & \,+ \, & (0.329)\\
            & | \, * & (0.342)\\
            & | \, - & (0.329)\\
{<}\text{var}{>} ::= &  \, \text{x} \, & (0.508) \\ 
            & | \, \text{1.0} \, & (0.493) \\
\end{align*}}}}

\caption{\label{fig:pge}Example of probability updating in PGE}
\end{figure}

We are going to use the example of grammar and individual presented Fig. \ref{fig:pcfg}, to show how the probabilities are updated. On the left panel of Fig. \ref{fig:pge} we can see the derivation tree of the individual that was used to update the probabilities of the PCFG on the right. The learning factor used was 0.01 (1\%).

Since the symbol $<start>$ has only one expansion rule, the probability of choosing $<expr>$ stays always 1. Looking at the derivation tree, we can see that the first derivation rule ($<expr><op><expr>$) of the non-terminal $<expr>$ was selected once, and the second derivation rule ($<var>$) was selected twice. 

Using the Alg. \ref{algo}, the new probabilities for the first derivation option is $min(0.5 + \frac{0.01*1}{3},1)$ equals 0.5033(3) , and for the second rule, $min(0.5 + \frac{0.01*2}{3},1)$ equals 0.5066(6). As the sum of the probabilities surpass 1, the excess ((1 - 0.5066 + 0.5033)/2 = 0.00495) is subtracted from both probabilities and the value is rounded to 3 decimal places. Then we distribute the excess for the derivations rules: the first rule is updated to 0.498 and the second rule is updated to 0.502.
This process is applied to the other symbols. For the non-terminal $<op>$ the rule $+$ and $-$ were never chosen so they will update equally ($(0.33 - 0.01*0.33)$ that equals 0.3267), and the rule $*$ was chosen once ($min(0.33 + \frac{0.01*1}{1},1)$ that equals 0.34). As the sum of the three probabilities is smaller than one, 0.0022 must be added to the three probabilities, and the result rounded, staying with 0.329 for the $+$ and $-$ symbols, and 0.342 for the $*$. 
The non-terminal $<var>$ was expanded twice for the terminal $x$ and never expanded for the terminal $1.0$. By applying the algorithm, the first rule ($x$) should be updated to 0.51 ($min(0.5+\frac{0.01*2}{2},1)$) and the second to 0.495 ($0.5-0.01*0.5$), being the sum of the two probabilities different than one, the adjustment and rounding should be done and they are updated to 0.508 and 0.493, respectively. 

\section{Experimental Analysis}
\label{sec:ea}

Over the years, several genotype-phenotype mapping alternatives have been proposed to increase the performance of GE. One that has obtained promising results is Structured Grammatical Evolution (SGE) \cite{Loureno2016,Loureno2017}. To evaluate the performance of PGE, we considered the standard GE and SGE algorithms in two different benchmark problems. These benchmark problems were selected taking into account the comparative analysis followed by \cite{Loureno2017} and the recommendations presented in \cite{McDermott2012}. For our experimental analysis we selected the Pagie Polynomial and the Boston Housing prediction problem.

\subsection{Problem Description}

\subsubsection{Pagie Polynomial}
Popular benchmark problem for testing Genetic Programming algorithms, with the objective of finding the mathematical expression for the following problem:

\begin{equation}
    \frac{1}{1+x^{-4}}+\frac{1}{1+y^{-4}} .
\end{equation}

The function is sampled in the interval [-5, 5.4] with a step of 0.4, and the grammar used is:

\setlength{\grammarindent}{6em} 
\begin{grammar}    

    <start> ::= <expr>
    
    <expr> ::= <expr> <op> <expr> | ( <expr> <op> <expr> )
    \alt <pre_op> ( <expr> ) | <var>
    
    <op> ::= + | - | * | /
    
    <pre\_op> ::= sin | cos | exp | log | inv
    
    <var> ::= x | y | 1.0
\end{grammar}

The division and logarithm functions are protected, i.e., $1/0 = 1$ and $log(f(x)) = 0$ $ if f(x) \le 0$.

\subsubsection{Boston Housing}
This is a famous Machine Learning problem to predict the prices of Boston Houses. The dataset comes from the StatLib Library \cite{BostonHousing} and has 506 entries, with 13 features. It was divided in 90\% for training and 10\% for testing. The grammar used for the Boston Housing regression problem is as follows:

\setlength{\grammarindent}{6em} 
\begin{grammar}    

    <start> ::= <expr>
    
    <expr> ::= <expr> <op> <expr> | ( <expr> <op> <expr> )
    \alt <pre_op> ( <expr> ) | <var>
    
    <op> ::= + | - | * | /
    
    <pre\_op> ::= sin | cos | exp | log | inv
    
    <var> ::= x[1] |...| x[13] | 1.0
\end{grammar}

\subsection{Parameters}
For all the experiments reported, the fitness function is computed using the Root Relative Squared Error (RRSE), where lower values indicate a better fitness. The parameters are presented in Table \ref{params}. These parameters were selected in order to make the comparisons between all the approaches fair. Additionally, and to avoid side effects, the wrapping mechanism was removed from GE and PGE. Concerning the variation operators for PGE, we used the standard one-point crossover, and float mutation which generates a new random float between [0,1]. Additionally, PGE uses a learning factor of $\lambda =1.0\%$.

\begin{table}
\caption{Parameters used in the experimental analysis for GE, PGE and SGE}
\centering
\begin{tabular}{lccc}
\hline
\multicolumn{1}{l}{} & \multicolumn{3}{c}{\textbf{Value}} \\ \hline
Parameters            & GE          & PGE       & SGE \\ \hline
Population Size       & \multicolumn{3}{c}{1000}      \\
Generations           & \multicolumn{3}{c}{50}        \\
Elitism               & \multicolumn{3}{c}{10\%}      \\
Mutation Probability  & \multicolumn{3}{c}{5\%}       \\
Crossover Probability & \multicolumn{3}{c}{90\%}      \\
Tournament            & \multicolumn{3}{c}{3}         \\
Max Number of Wraps   & \multicolumn{2}{c}{0}   & -   \\
Size of Genotype      & \multicolumn{2}{c}{128} & -   \\
Max. Initialisation depth        & \multicolumn{2}{c}{-}   & 6   \\
Max. Tree Depth        & \multicolumn{2}{c}{-}   & 17   \\ \hline
\end{tabular}
\label{params}
\end{table}

\section{Results} 
\label{sec:results}
The experimental results in this section will be presented in terms of the mean best fitness, which results from the execution of 100 independent runs. To compare all approaches we performed a statistical study. Since the results did not follow any distribution, and the populations were independently initialised, we employed the Kruskal-Wallis non-parametric test to check if there were meaningful differences between the different groups of approaches. When this happened we used the Mann-Whitney \textit{post-hoc} test with Bonferroni correction. For all the statistical tests we considered a significance level $\alpha$ = 0.05. 

\begin{figure}[ht]
    \centering
    \includegraphics[height=6cm]{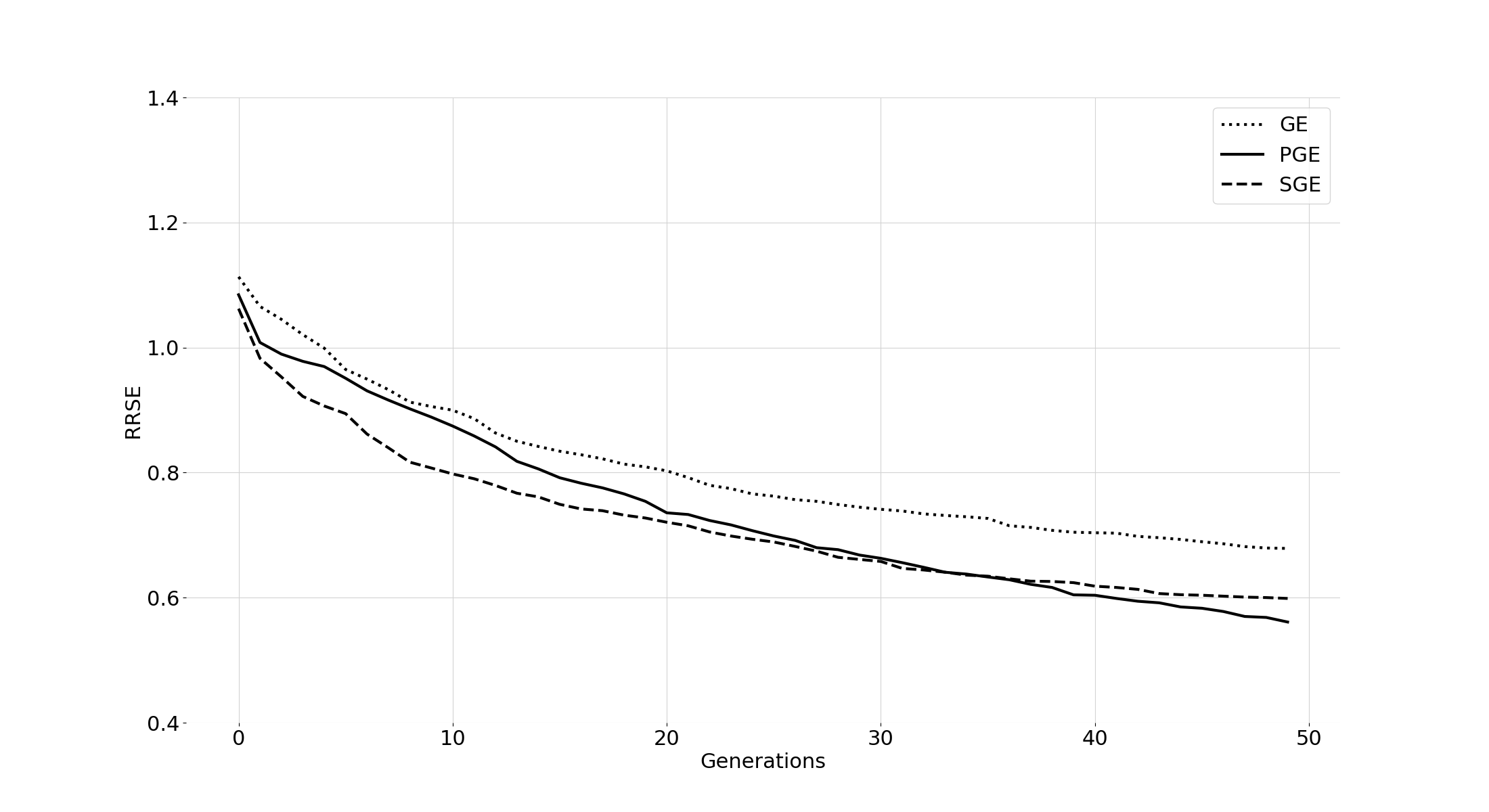}
    \caption{Results for the Pagie Polynomial}
    \label{fig:pagie}
\end{figure}

Fig. \ref{fig:pagie} depicts the results for the Pagie Polynomial. It is possible to see that all the methods start from similar fitness values, but as the evolutionary process progresses, differences between the approaches emerge. The fitness of the solutions being evolved by SGE rapidly decrease in the early generations, but slows down after a certain number of generations ($\approx$ 20). For GE, the fitness decreases slowly through the generations. This results are in line with the ones presented in \cite{Loureno2016}.

The PGE performance is different from both SGE and GE. In the early generations PGE decreases slowly (following the same trend as GE), but then, around the 10th generation, the fitness of PGE starts to rapidly decrease. Around generation 35 it surpasses the quality of SGE. The first row of Table \ref{std} shows the Mean Best Fitness and the Standard Deviation for each approach. We can see that for the Pagie Polynomial problem PGE obtains the lowest error. 

\begin{table}
\centering
\caption{Mean Best Fitness and Standard Deviation for all the methods used in the comparison. Results are averages of 100 independent runs.}
\begin{tabular}{lccc}
\hline
Problem     & \multicolumn{1}{l}{PGE}     & \multicolumn{1}{l}{GE}   & \multicolumn{1}{l}{SGE}   \\ \hline
Pagie Polynomial        & \textbf{0.56$\pm$0.16} & 0.68$\pm$0.17 & 0.59$\pm$0.13 \\
Boston Housing Train & 0.82$\pm$0.12          & 0.88$\pm$0.14 & \textbf{0.78$\pm$0.13}  \\
Boston Housing Test     & 0.84$\pm$0.13 & 0.90$\pm$0.15 & \textbf{0.79$\pm$0.12}     \\ \hline

\end{tabular}
\label{std}
\end{table}
In terms of statistical significant differences, the Kruskall-Wallis showed meaningful differences between the approaches. The \textit{post-hoc} results are depicted in Table \ref{stats}. Looking at the results it is possible to see that both PGE and SGE are better than GE with statistical significant differences. When comparing PGE and SGE we only found marginal differences (p-value = 0.04) on the Boston Housing Training.


\begin{table}
\centering
\caption{Results of the Mann-Whitney \textit{post-hoc} statistical tests. Bonferroni correction is used and the significance level $\alpha=0.05$ is considered.}
\begin{tabular}{lcc}
\hline
                        & \multicolumn{1}{l}{PGE - GE} & \multicolumn{1}{l}{PGE - SGE} \\ \hline
Pagie Polynomial        & \textbf{0.00}                & 0.24                          \\
Boston Housing Training & \textbf{0.00}                & 0.04                          \\
Boston Housing Test     & \textbf{0.00}                & 0.10  \\ \hline
\end{tabular}
\label{stats}
\end{table}

\begin{figure}[ht]
\begin{subfigure}{0.5\textwidth}
    \centering
    \includegraphics[height=3.7cm]{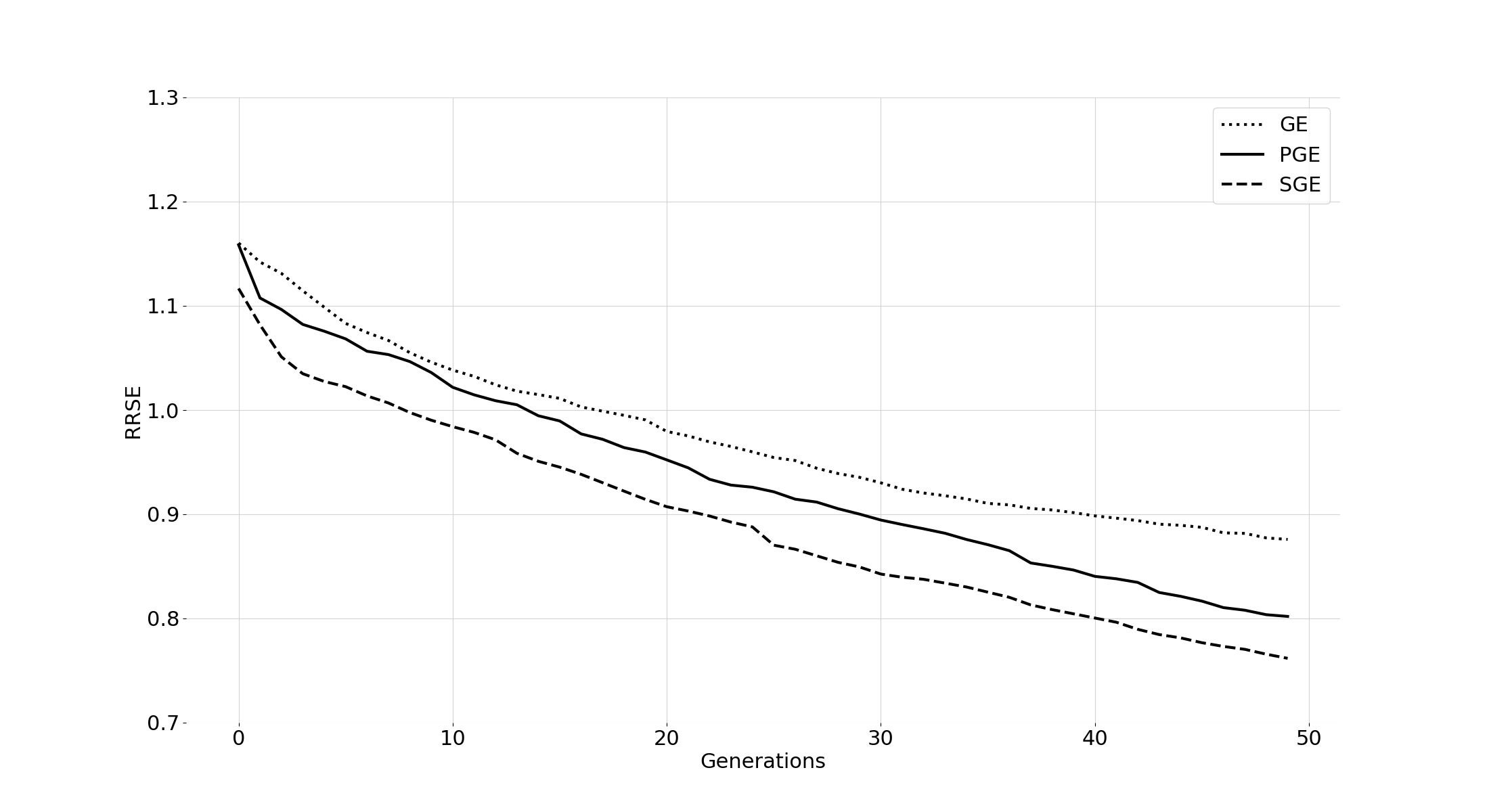}
    \caption{Training}
\end{subfigure}
\begin{subfigure}{0.5\textwidth}
    \centering
    \includegraphics[height=3.7cm]{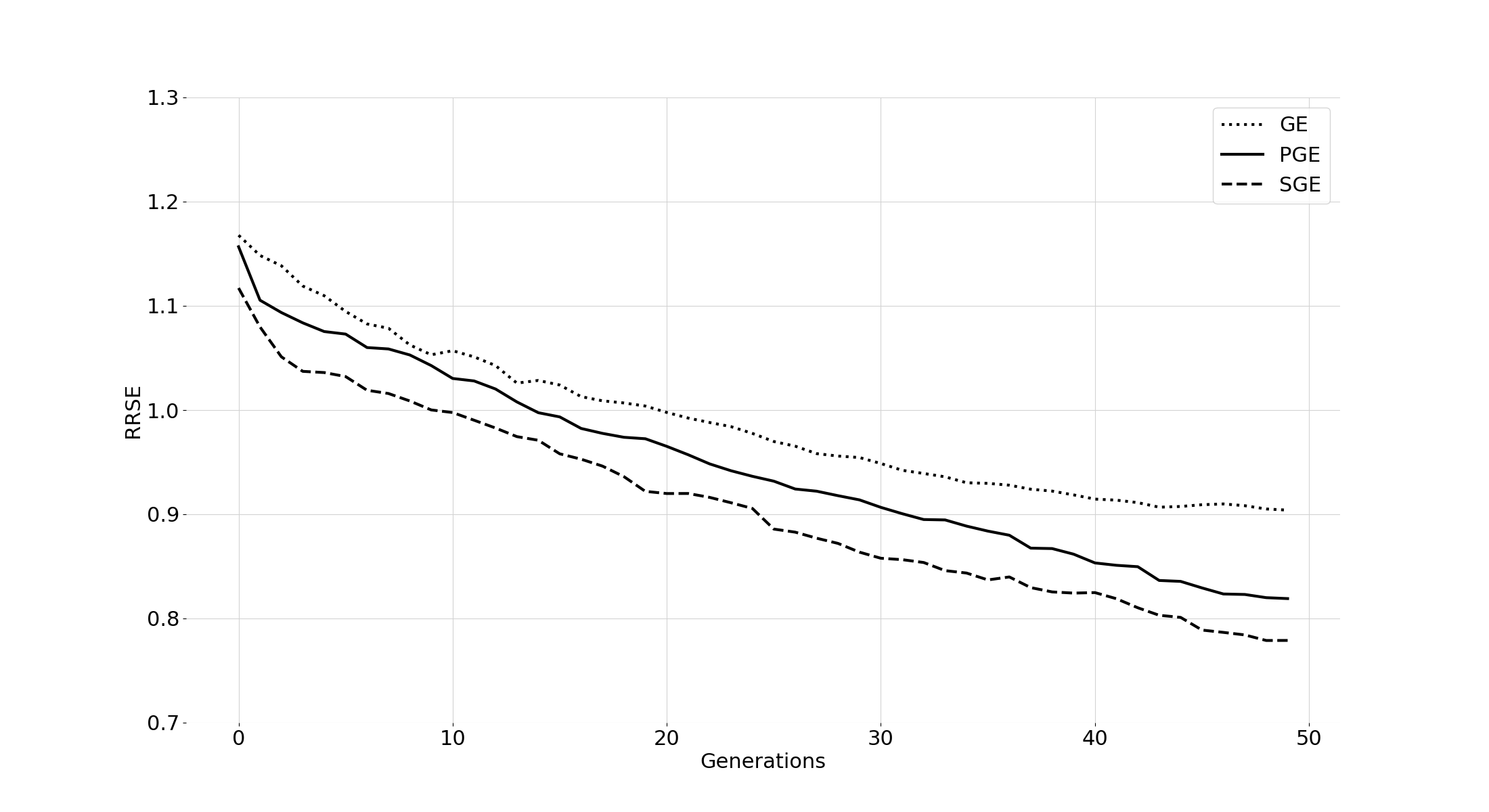}
    \caption{Testing}
\end{subfigure}
\caption{Results for the Boston Housing problem}
    \label{fig:bh}
\end{figure}

Fig. \ref{fig:bh} shows the results for the Boston Housing problem. Looking at the training results (Fig. \ref{fig:bh} (a)), it is possible to see that the fitness of SGE individuals rapidly decrease, and continue too over the entire evolutionary process. The performance of GE is in line with what we observed previously, i.e.,  a slow decrease on the fitness. Even though the training results are important to understand the behaviour of the methods, the testing results are more relevant, because they allow us to evaluate the generalisation ability of the models evolved by each approach. Looking at the test results (Fig. \ref{fig:bh}(b)) we can see that SGE and PGE are building models that can generalise better to unseen data.


Once again we applied a statistical analysis to check whether there were differences between the approaches (Table \ref{stats}). The results, for both training and test, show that PGE is statistically significant than GE, but there are no differences between PGE and SGE.

Finally we present an analysis on how the probabilities of certain derivation rules progress over the generations. This analysis will give us insights into what are the rules that are more relevant. 

\begin{figure}[ht]
    \centering
    \includegraphics[height=6cm]{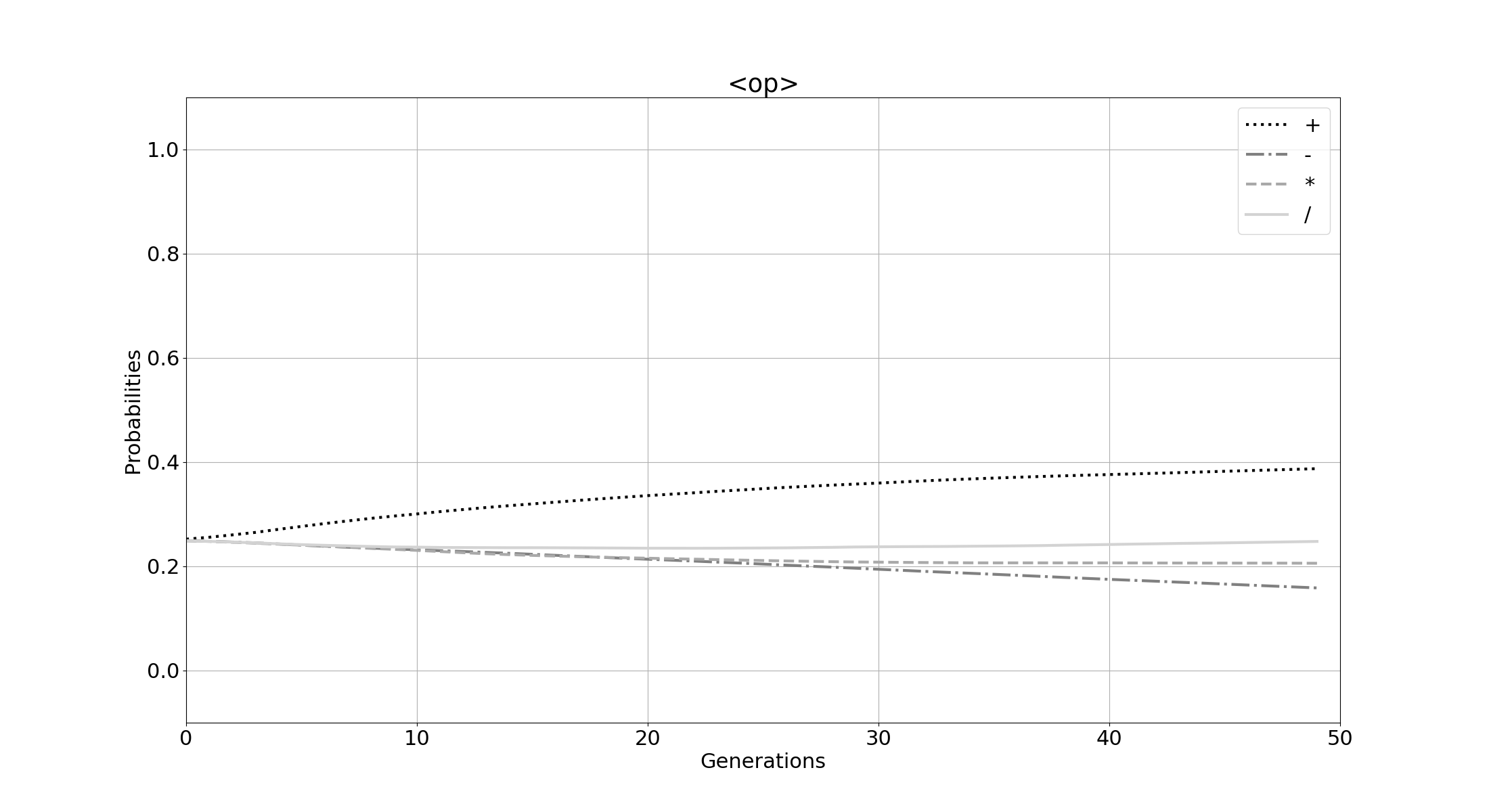}
    \caption{Evolution of grammar probabilities of non-terminal $<op>$ with the Pagie Polynomial. Results are averages of 100 runs.}
    \label{fig:probspagie}
\end{figure}

For the Pagie Polynomial, Fig. \ref{fig:probspagie} presents the evolution of the PCFG's probabilities for the non-terminal $<op>$ over the generations. As one would expect, the probabilities associated with the symbols that are required to solve the problem are higher, namely the ones associated with the terminal symbols $+$ and $/$.

\begin{figure}[ht]
    \centering
    \includegraphics[height=6cm]{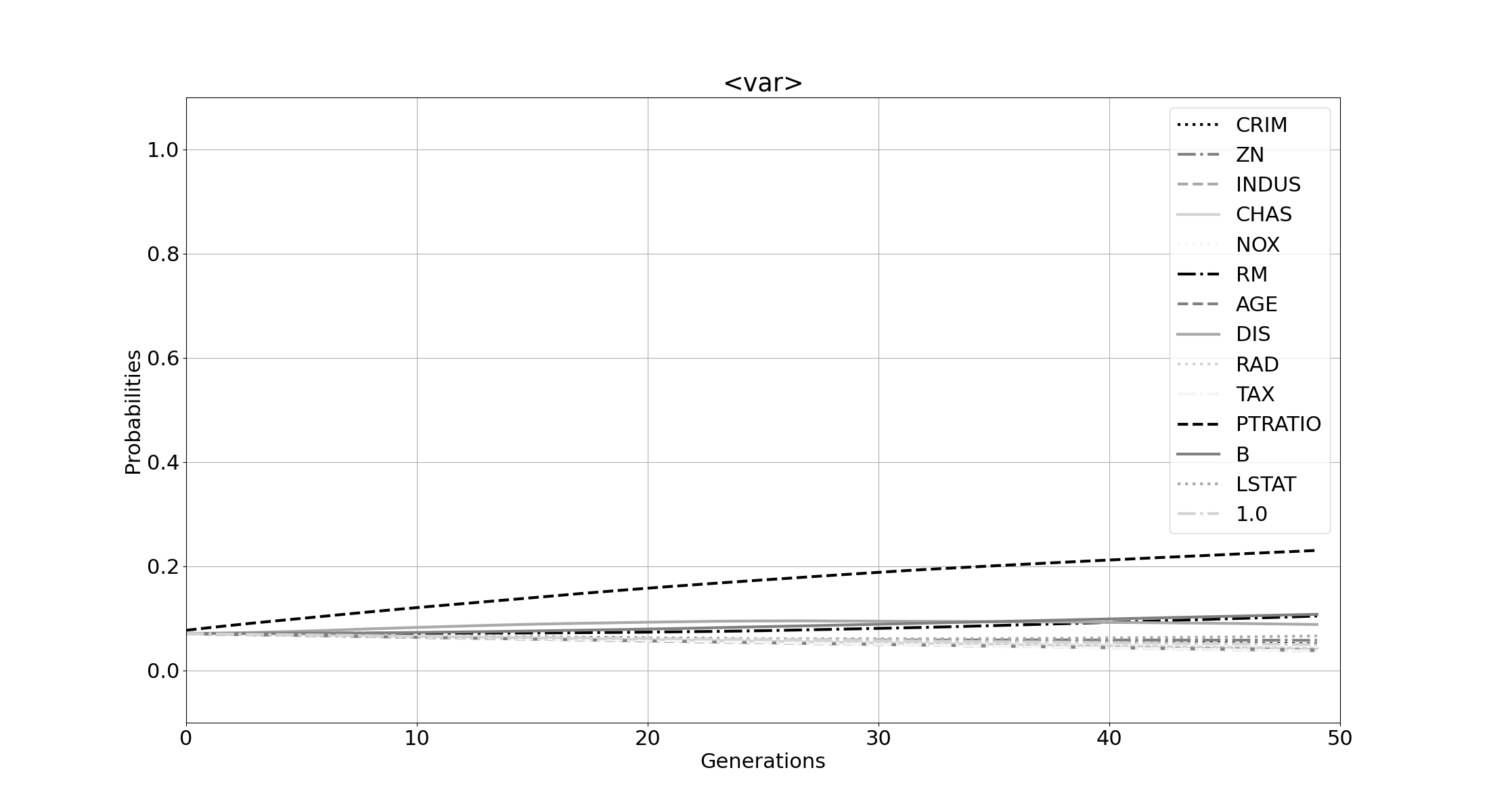}
    \caption{Evolution of grammar probabilities of non-terminal $<var>$ with the Boston Housing problem. Results are averages of 100 runs.}
    \label{fig:probsbh}
\end{figure}

Concerning the Boston Housing, the progression of the probabilities are depicted in Fig. \ref{fig:probsbh} and in Table \ref{probsbhtable}. Concretely, the results show the probabilities for the derivation options of the non-terminal $<var>$. This symbol was selected because it contains the features that describe the problem. Looking at the evolution of the probabilities associated with each production, we can understand which of these features are more relevant to accurately predict the price of houses. Looking at the results (Fig. \ref{fig:probsbh}), one can see that PTRATIO stands out in terms of the probability of being selected. PTRATIO represents the pupil-teacher ratio by town. This is in line with the results reported by \cite{Che2017}. Another interesting result is to see that the feature RM (the third most important feature in \cite{Che2017}), which is the average number of rooms per dwelling, is also on the top three of our results. These results confirm not only the relevance of these features to the Boston Housing problem, but also allow us to perform feature selection and provide an explanation to the results achieved. This means that at the end of the evolutionary process one can look at the final distribution of the probabilities in the grammar, and analyse the relative importance of each production and derivation rules, and see how they are used to create the best models.

\begin{table}
\centering
\caption{Probabilities of the Boston Housing Dataset's productions of the non-terminal $<var>$ at the end of the evolutionary process. Results are averages of 100 runs.}
\begin{tabular}{lr}
\hline
Production                  & \multicolumn{1}{l}{Probability} \\ \hline
PTRATIO                     & 0.23                            \\
B                           & 0.11                            \\
RM                          & 0.1                             \\
DIS                         & 0.09                            \\
LSTAT                       & 0.07                            \\
ZN                          & 0.06                            \\
NOX, CRIM, 1.0              & 0.05                            \\
RAD, TAX, RADIUS, CHAS, AGE & 0.04                            \\ \hline
\end{tabular}
\label{probsbhtable}
\end{table}

\section{Conclusion}
\label{sec:conclusion}

Grammatical Evolution (GE) has attracted the attention of many researchers and practitioners. Since its proposal in the late 1990s, it has been applied with success to many problem domains. However, it has been shown that it suffers from some issues.  

In this paper we proposed a new mapping mechanism and genotypic representation called Probabilistic Grammatical Evolution (PGE). In concrete, in PGE the genotype of an individual is a variable length sequence of floats, and the genotype-phenotype mapping is performed using a Probabilistic Context-Free Grammar (PCFG). Each derivation rule has a probability associated, and they are updated with taking into account the number of times that the derivation rules were selected by the best individuals. In order to maintain a balance between global and local exploration we alternate between the best overall individual and the best individual of generation, respectively. 

PGE was compared with standard GE and SGE in two different benchmarks. The results show that for both problems PGE is statistically better that GE and has a similar performance when compared to SGE. We also analyse how the probabilities associated the different productions progress over time, and it was possible to see that the production rules that are more relevant to the problem at hand have higher probabilities of being selected.

In terms of future work one needs to consider alternative mechanisms to adjust probabilities of the productions rules. Another line of work that needs to be conducted is concerned with the analysis of the locality and redundancy in PGE.

\section*{Acknowledgements}
This work is partially funded by the project grant DSAIPA/DS/0022/2018 (GADgET), by national funds through the FCT - Foundation for Science and Technology, I.P., within the scope of the project CISUC - UID/CEC/00326/2020 and by European Social Fund, through the Regional Operational Program Centro 2020. We also thank the NVIDIA Corporation for the hardware granted to this research.

\bibliography{references}
\bibliographystyle{splncs03}

\end{document}